# Resource-Aware Programming for Robotic Vision

Johny Paul, Walter Stechele
Institute for Integrated Systems
Technical University of Munich, Germany
{Johny.Paul, Walter.Stechele}@tum.de

Manfred Kröhnert, Tamim Asfour
Institute for Anthropomatics
Karlsruhe Institute of Technology, Germany
{Kroehnert, Asfour}@kit.edu

*Abstract*— Humanoid robots are designed to operate in human centered environments. They face changing, dynamic environments in which they need to fulfill a multitude of challenging tasks. Such tasks differ in complexity, resource requirements, and execution time. Latest computer architectures of humanoid robots consist of several industrial PCs containing single- or dual-core processors. According to the SIA roadmap for semiconductors, many-core chips with hundreds to thousands of cores are expected to be available in the next decade. Utilizing the full power of a chip with huge amounts of resources requires new computing paradigms and methodologies.

In this paper, we analyze a resource-aware computing methodology named Invasive Computing, to address these challenges. The benefits and limitations of the new programming model is analyzed using two widely used computer vision algorithms, the Harris Corner detector and SIFT (Scale Invariant Feature Transform) feature matching. The result indicate that the new programming model together with the extensions within the application layer, makes them highly adaptable; leading to better quality in the results obtained.

## I. INTRODUCTION

Humanoid robots have to fulfill many different tasks during operation including recognizing objects, navigating towards a goal region, avoiding collisions, maintaining balance, planning motions, having conversations, and many more. Additionally they have to work in highly dynamic and ever changing human centered environments. Taking a closer look, all of these tasks have different requirements depending on their objectives. Some tasks are computationally or data-flow intensive and some are control flow intensive. Some are continuously running at a specific frequency while others are doing their work asynchronously triggered by external events.

### A. Conventional Architectures used on Humanoid Robots

Current humanoid robot control architectures share similar hardware setups consisting of one or more industrial PCs equipped with single- or dual-core processors. These architectures are quasi-parallel in the sense that the applications are mapped statically to various compute units. The architecture of the humanoid robot Asimo uses two PCs, a control and planning processor plus an additional digital signal processor (DSP) for sound processing [15]. Similarly, two processor boards (one for realtime-control, one for non-realtime tasks) are used by the humanoid robots HRP-2 and HRP-4C [8], [9]. In contrast, there are five industrial PCs and PC/104 systems built into the humanoid robot ARMAR-III (Fig. 1) each dedicated to different tasks like computer vision, low-level control, high-level control, and speech processing [1].

All presented robots are autonomous as they do not rely on external resources while operating. Most systems use standard PC components for computation while sometimes utilizing dedicated hardware for specialized tasks such as DSPs for sound recognition. On these mostly homogeneous architectures almost all tasks share the same set of resources. However, one PC is sometimes dedicated to speech recognition and synthesis.

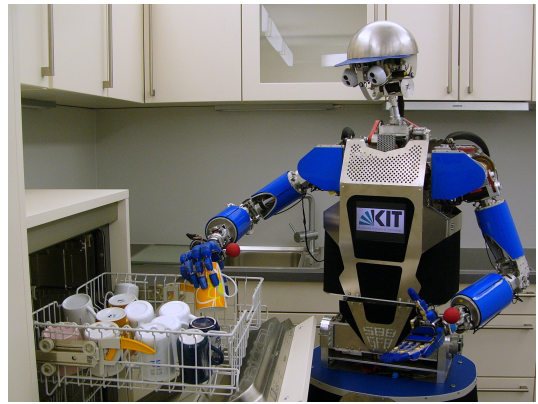

Fig. 1. Humanoid Robot ARMAR-III

### B. Challenges and difficulties on Existing Hardware

Robot-specific tasks need to be executed on the hardware built into the robot. Timely execution of tasks is guaranteed as the underlying hardware is selected to be able to cope with worst case requirements.

In the past, Graphics Processing Units (GPU) and Field Programmable Gate Arrays (FPGA) have offered significant speedup to computer vision applications. In [3], [10], various case studies have been presented comparing GPU vs. FPGA implementations. There seems to be no "one fits all" solution, but the benefits and drawbacks are manifold: GPUs benefit from higher clock frequencies than FPGAs, due to their custom layout. On the other hand, programming GPUs efficiently can be as tedious as hardware development in VHDL and their massive power consumption prevent them from being used on mobile robots. FPGAs offer a higher degree of freedom in optimizing memory architectures, data types and pipeline structures at lower power consumption.

However, programming FPGAs require HDL experience, and FPGA designs consume more time during development and testing.

The use of many-core processors can mitigate some of the above mentioned problems on account of their immense computational power assembled in a compact design. Embracing emerging many-core technologies like Intels Single-Chip Cloud Computer (SCC) [7] or Tileras Tile64 [2] seems evident. Once many-core technologies become a de-facto standard the question arises of how to handle such large amounts of resources and how to program these systems efficiently. A common practice as of now is to do static resource allocation at compile time as described in [12]. However, static allocation schemes have issues with dynamic scenarios. Often, the changing requirements lead to under-utilization since resources are often occupied without performing any useful work.

On the other hand, dynamic scheduling can improve resource utilization. However, the available resources on a many-core chip (processing elements (PEs), memories, interconnects, etc.) have to be shared among various applications running concurrently, which leads to unpredictable execution time or frame drops for vision applications.

This paper is organized as follows. Section II describes some of the challenges faced by vision applications on a conventional many-core system using the example of a widely used image processing algorithm called Harris Corner detector. In Section III we introduce the concepts of resource-aware computing including proposed hardware units. Section IV presents our evaluation and results from the resource-aware model using two different applications, Harris Corner and SIFT (Scale Invariant Feature Transform) feature matching. Finally Section V concludes the paper.

## II. PROBLEM DESCRIPTION

Corner detection is often employed as the first step in computer-vision applications with real-time video input. Hence, the application has to maintain a steady throughput and good response time to ensure quality results. However, the presence of other high-priority tasks may alter the behavior of the corner-detection algorithm. To evaluate such a dynamically changing situation, we analyzed the behavior of the conventional Harris detector on a many-core processor with 32 PEs. A video input with $640 \times 480$ pixels at 10 frames per second was used, with the test running for 20 seconds. To evaluate the impact of other applications running concurrently on the many-core system, applications like audio processing, motor control, etc. were used. These applications create dynamically changing load on the processor based on what the robot is doing at that point in time. For instance, the speech-recognition application is activated when the user speaks to the robot.

Sharing of available resources among various applications running concurrently, resulted in a resource allocation pattern as shown in Fig. 2 and an execution-time profile shown in Fig. 3. It can be seen that the execution time varies from 0 to 430 milliseconds, based on the load condition. A lack of sufficient resources leads to very high processing intervals or frame drops (a processing interval of zero represents a frame drop). The number of frames dropped during this evaluation is as high as **20%** and the worst-case latency increased by 4.3x (100 milliseconds to 430 milliseconds). Frame drops reduce the quality of the results and the robot may lose track of the object if too many consecutive frames are dropped.

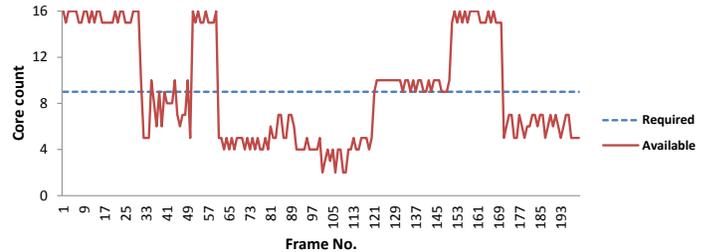

Fig. 2. Resource allocation scheme

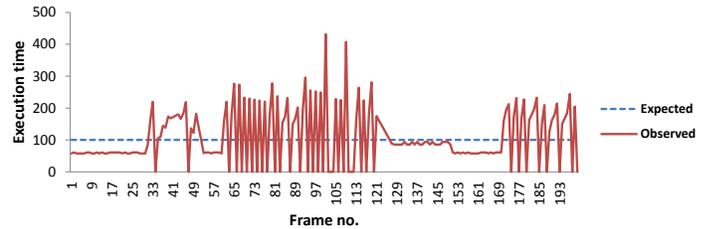

Fig. 3. Variation in processing interval based on available resources

We attempt to tackle some of these issues though a novel resource-aware programming methodology called Invasive Computing. We envision that the use of Invasive Computing mechanisms into future computing paradigms will significantly contribute to solve this problem. This work also describes how to distribute the huge workload on the massively parallel PEs for best performance, and how to generate results on time (avoiding frame drops) even under varying load conditions.

## III. INVASIVE COMPUTING

Invasive Computing as a synonym for resource-aware computing was first introduced in 2008 [17]. One major goal of the project is to investigate more elegant and efficient ways of dealing with massive parallelism within the future many-core platforms. In this section we describe the nature and features of an Invasive Computing platform and outline the differences and advantages compared to other available platforms.

### A. Programming Model for Invasive Computing

Invasive Computing is about managing parallelism in and between programs on a Multi-Processor System on Chip (MPSoC). This is achieved by empowering applications to manage processing-, memory-, or communication-resources themselves, hence the term resource-aware computing. Resource management in an invasive system is divided into three distinct phases, also depicted in Fig. 4:

1) *invade*: acquire resources
2) *infect*: use resources
3) *retreat*: release resources

Programs can specify amounts of resources needed for execution in the *invade*-phase. Granted resources are used in the *infect*-phase for example by distributing program binaries onto processing elements. Resources can be released after usage in the *retreat*-phase. However, it is also possible to keep acquired resources and directly proceed with a new *infect*-phase.

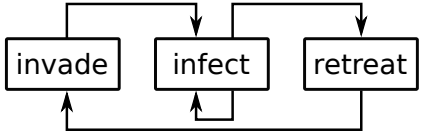

Fig. 4. Phases of Invasive Computing

In contrast to static methods the invasive approach leads to a more decentralized resource allocation scheme. This allows programs to self-explore the system they are running on by requesting and releasing resources depending on the current degree of parallelism. Granted resources may differ from requested ones, a situation where the affected programs must adapt to. In case of fewer assigned resources a program might choose a different algorithm in order to proceed. This self-adaption to current resource distributions can lead to increased overall system stability and failure tolerance as well as better load balancing.

### B. Hardware Architecture for Invasive Computing

Our target many-core processor has a tiled architecture interconnected by a NoC [6], as shown in Fig. 5. Each compute tile consists of 4 cores interconnected by a local bus and some fast, on-chip tile-local memory, with a total of 32 cores (LEON3, a SPARC V8 design by Gaisler [4]) spread across 8 tiles. The 9th tile is a memory and I/O tile encompassing a DDR-III memory controller and Ethernet, UART, etc. for data exchange and debugging. Each core has a dedicated L1 cache while all the cores within a tile share a common L2 cache for accesses that go beyond the tile boundary to the external DDR-III memory. L1 caches are write-through and L2 is a write-back cache. Cache coherency is only maintained within the tile boundary to eliminate a possible scalability bottleneck when scaling to higher core counts. Therefore, data consistency has to be handled by the programmer through proper programming techniques that are built on top of hardware features to provide consistent data access and exchange between the different cache-coherency domains. This scheme is somewhat similar to the Intel SCC. The prototype design runs on a multi-FPGA platform from Synopsys called CHIPit System [16] (consisting of six Xilinx Virtex-5 XC5VLX330 FPGAs) with the clock frequency set to 50 MHz.

## IV. RESOURCE-AWARE APPLICATIONS

This section describes the resource-aware model for Harris Corner and SIFT feature matching based on KD-Trees and also demonstrates the benefits obtained through the resource-aware programming model. The results obtained are compared with their conventional counterparts.

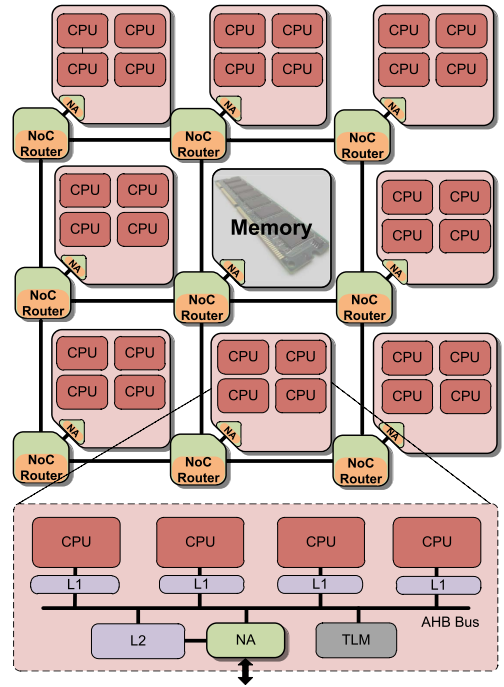

Fig. 5. InvasIC Hardware Architecture

### A. Harris Corner Detection Algorithm

The humanoid robot ARMAR-III uses a Harris Corner detector [5] as the first stage in the object-recognition and -tracking algorithm and has to operate on the real-time video stream produced by the cameras on the robot. Hence, it has to maintain a steady throughput and good response times to ensure high-grade results, because any deterioration in the quality of the detected corners will negatively affect the remaining stages of the object-recognition algorithm. The quality loss encountered by Harris Corner detector due to frame drops has already been discussed in Section II. The use of the conventional algorithm resulted in very high latencies under circumstances where sufficient resources are not available, and dropped frames occasionally.

The resource-aware model of Harris Corner [14] is based on the idea that in most situations, the obvious non-corners constitute a large majority of the image. Hence the Harris detectors incur a lot of redundant computations as they evaluate the entire image for a high corner response. The conventional Harris Corner detector is based on the local auto-correlation function that is approximated by a matrix $M$ over a small window $w$ for each pixel $p(x,y)$:

$$M = \left[ \begin{array}{cc} \sum_w W(x)I_x^2 & \sum_w W(x)I_xI_y \\ \sum_w W(x)I_xI_y & \sum_w W(x)I_y^2 \end{array} \right] = \left[ \begin{array}{cc} a & b \\ c & d \end{array} \right] \quad (1)$$

where $I_x$ and $I_y$ are horizontal and vertical intensity gradients, respectively, and $W(x)$ is an averaging filter that can be a box or a Gaussian filter. The eigenvalues $\lambda_1$ and $\lambda_2$ (where $\lambda_1 \geq \lambda_2$) indicate the type of intensity change in the window $w$ around $p(x, y)$. If both $\lambda_1$ and $\lambda_2$ are small, $p(x, y)$ is a point in a flat region. If $\lambda_1$ is large and $\lambda_2$ is small, $p(x, y)$ is an edge point and if both $\lambda_1$ and $\lambda_2$ are large, $p(x, y)$ represents a corner point. Harris combines the eigenvalues into a single corner measure $R$ as shown in Equation 2 ($k$ is an empirical constant with value 0.04 to 0.06). Once the corner measure is computed for every pixel, a threshold is applied on the corner measures to discard the obvious non-corners.

$$R = \lambda_1 \lambda_2 - k \cdot (\lambda_1 + \lambda_2)^2 = (ac - b^2) - k \cdot (a + c)^2 \quad (2)$$

In order to enhance the conventional detector to a resource-aware detector, a corner response ($CR$) is defined in Equation 3, where product of vertical and horizontal difference in pixel intensities is used and the candidates with low $CR$ values are pruned away.

$$CR = (|I_x \cdot I_y|) \quad (3)$$

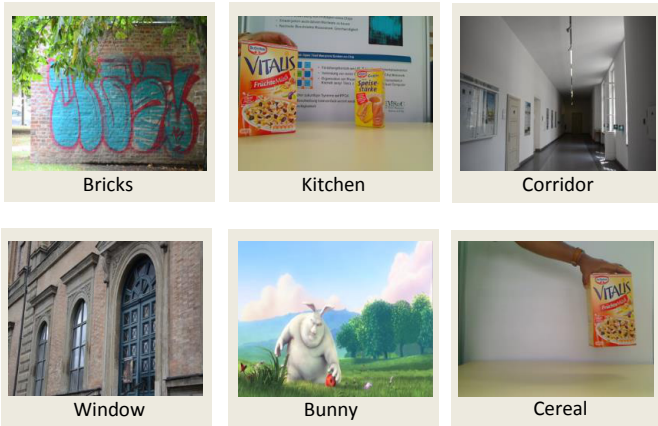

Fig. 6. Snapshot of the video-sequences used for evaluation

In order to evaluate the impact of pruning on the accuracy of detected corners, we use the metrics named *precision* and *recall* as proposed in [11]. The value of recall measures the number of correct matches out of the total number of possible matches, and the value of precision measures the number of correct matches out of all matches returned by the algorithm. As an aftereffect of pruning, the values of precision and recall drops slightly in the region where the application has to adapt by pruning pixels. However, this helps to avoid overshoot in execution time and eliminate frame drops, so that results are consistently available within the predefined intervals. An overall comparison between the two scenarios is shown in Table I.

The use of the conventional algorithm leads to a very high worst-case execution time(WCET) and frame drops. The precision and recall values are low for the conventional algorithm as a frame drop leads to zero precision and recall for that particular frame. In brief, the resource-aware Harris detector can operate very well under dynamically changing conditions by adapting the workload, avoiding frame drops and regulating the WCET, leading to high precision and recall rates.

|  | Throughput | WCET | Precision | Recall |
|---|---|---|---|---|
| Conventional | **81%** | 4.31x | 0.82 | 0.81 |
| Resource-aware | 100% | 1.04x | **0.98** | **0.98** |

TABLE I
COMPARISON BETWEEN CONVENTIONAL AND RESOURCE-AWARE HARRIS DETECTORS

It is interesting to note that the effects of pruning vary based on the scene. For example, the speedup achieved (using the same threshold) is low for cluttered scenes like *Bricks* while the majority of the pixels can be pruned away for scenes with plain backgrounds. This means that the amount of computing resources required to perform the corner detection will vary from one scene to another based on the nature of the foreground, background, etc. and therefore the resources have to be allocated on a frame-to-frame basis, based on the scene captured.

### B. Nearest-Neighbor Search on Kd-Trees

The object-recognition process used on the ARMAR robot consists of two major steps. In the first step, the robot is trained to recognize the object. A training data set consisting of SIFT features is created for every known object to be recognized. The second step in the recognition process has real-time requirements as it helps the robot to interact with its surroundings (by recognizing and localizing various objects) in a continuous fashion. In this step, a set of SIFT features, extracted from the real-time images, is compared with the training data set using a nearest neighbor (NN) search. The computation of the nearest neighbor for the purpose of feature matching is the most time-consuming part of the complete recognition and localization algorithm. This algorithm performs a heuristic search and only visits a fixed number of leaves resulting in an actual nearest neighbor, or a data point close to it.

For the NN-search algorithm, the number of kd-tree leaves visited during the search process determines the overall quality of the search process. Visiting more leaf nodes during the search leads to a higher execution time. The search duration per SIFT feature can be calculated from Fig. 7. The values were captured by running the NN-search application on a single PE using a library of input images covering various situations encountered by the robot.

From this graph it is clear that the search interval varies linearly with the number of leaf nodes visited during the search. Moreover, the relation between quality (i.e. the number of features recognized) and leaf nodes is shown in Fig. 8. The quality of detection falls rapidly when the number of leaf nodes is reduced below 20 and increases almost linearly in the

11

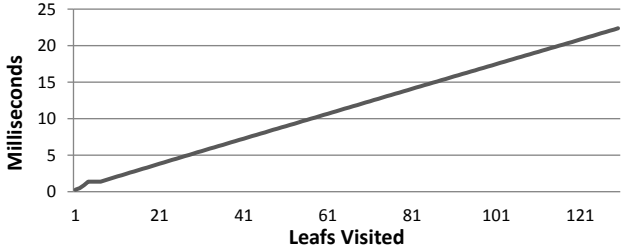

Fig. 7. Variation of execution time vs. leaf nodes visited for NN-search

range between 20 and 120.

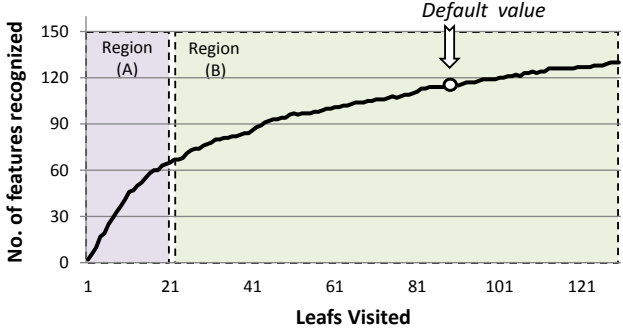

Fig. 8. Search quality vs. leaf nodes visited for NN-search

At a further higher leaf count, the quality does not improve significantly as all the possible features are already recognized. In the conventional algorithm used on CPUs, the number of leaf nodes visited is set statically such that the search process delivers results with sufficient quality for the specific application scenario. Using the results from this evaluation, the overall search duration can be predicted based on the object to be recognized, the number of features to be processed and the number of PEs available for NN-search. The first two parameters are decided by the application scenario while the PE count is decided by the runtime system based on the current load situation. Hence, the resources allocated to the application may vary from time to time, leading to highly unpredictable search durations and frame drops.

We try to address these issues using a resource-aware model for KD-Tree search [13]. The main idea and novelty of the resource-aware algorithm is that the workload is calculated and distributed taking into account the available resources (PEs) on the many-core processor. The amount of PEs requested by the algorithm is based on the number of SIFT features to be processed, the size of the kd-tree and the available search interval. The number of SIFT features varies from frame to frame based on the nature and number of objects present in the frame, the nature of the background, etc. The size of the kd-tree is decided by the texture pattern on the object to be recognized and tracked. The search interval or the frame rate is decided by the context where the NN-search is employed. For example, if the robot wants to track a fast-moving object, the frame rate has to be increased or the execution time has to be reduced. Equation 4 represents this relation and can be used to compute the number of PEs ($N_{pe}$) required to perform the NN-search on any frame within the specified interval $T_{search}$. $N_{fp}$ is the number of SIFT features to be processed and $T_{fp}$ is the search duration per SIFT feature, a function of the number of leaf nodes visited, as described in Fig. 7. The initial resource estimate is based on the default leaf count ($N_{leaf\_best}$), a statically defined value based on the application scenario.

$$N_{pe} \geq \frac{N_{fp} \times T_{fp}(N_{leaf\_best})}{T_{search}} \qquad (4)$$

Note that the function $T_{fp}(N_{leaf\_best})$ is different for every object to be recognized and tracked by the robot, as this is dependent on the number of features forming the kd-tree, the shape of the tree, etc. Using this model, the application raises a request to allocate PEs ($N_{pe}$), which is then processed by the operating system. Considering the current system load, the OS makes a final decision on the number of PEs to be allocated to the NN-search algorithm. The PE count may vary from zero (if the system is too heavily loaded and no further resources can be allocated at that point in time) to the total number of PEs requested (provided that there exists a sufficient number of idle PEs in the system and the current power mode offers sufficient power budget to enable the selected PEs). This means that under numerous circumstances the application may end up with fewer PEs and has to adapt itself to the limited resources offered by the runtime system. This is achieved by recalculating the number of leaf nodes ($N_{leaf\_adap}$) to be visited during the NN-search such that the condition in Equation 5 is satisfied.

$$T_{fp}(N_{leaf\_adap}) \leq \frac{N_{pe} \times T_{search} \times \eta(N_{pe})}{N_{fp}} \qquad (5)$$

The algorithm can use the new leaf count for the entire search process on the current image. It should be noted that the resource-allocation process operates once for every frame. Upon completion, the application releases the resources and waits for the next frame to arrive.

A set of 100 different scenes was used for evaluation, where each frame contains the object to be recognized and localized along with few other objects and changing backgrounds. The position of the objects and their distance from the robot were varied from frame to frame to cover different possible scenarios. Evaluations were conducted on the FPGA-based HW prototype, as described in Fig. 5. Fig. 9 shows a comparison between the resource-aware and the conventional NN-search, with the number of features recognized (quality of detection) on the y-axis and the frame number on the x-axis.

In order to maintain equality in the evaluation process, the number of PEs allocated to the applications was equalized. It is clear from Fig. 9 that the resource-aware NN-search algorithm outperforms the conventional algorithm using the same amount of resources, as the resource-aware model is capable of adapting the search algorithm based on the available resources compared to the conventional algorithm with fixed thresholds. However, the resource-aware algorithm results in the same number of matched features as the conventional algorithm in some frames. This is because there were a sufficient number of idle PEs and the runtime system allocated sufficient resources to meet the computing requirements of the conventional algorithm and hence the conventional algorithm did not drop any SIFT feature. On the contrary, when a frame contains large number of SIFT features and the processing system is heavily loaded by other applications, the conventional algorithm dropped too many SIFT features, thereby resulting in a low overall detection rate (matched features).

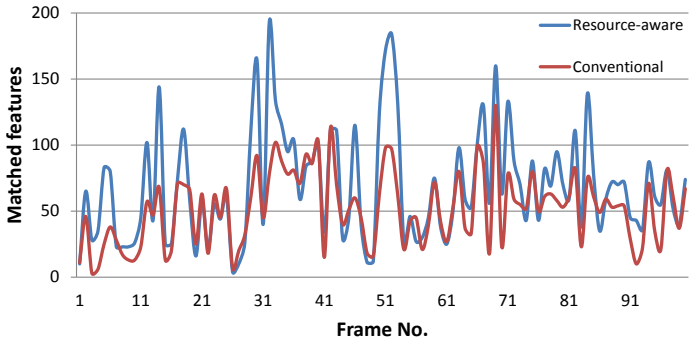

Fig. 9. Comparison between resource-aware and conventional NN-search

## V. CONCLUSION

This paper presented the benefits of the resource-aware programming model named Invasive Computing for robotic vision applications. Resource-aware Harris Corner detector and SIFT feature matching based on KD-Trees were presented, together with techniques on how to estimate the resources required for the computation based on the scene, the resolution of the input image and the user-specified time interval. The applications with resource-aware extension are aware of the availability of resources on the many-core processor and can adapt the workload if sufficient resources are not available. The enhanced corner detector and KD-tree search algorithms can generate results within the specified search interval and avoid frame drops. Our experiments show that incorporating resource awareness into the conventional vision algorithms can significantly improve their quality. A detailed evaluation was conducted on an FPGA-based hardware prototype to ensure the validity of the results. The resource allocation and release happens once per frame and the additional overhead in execution time is negligible when compared to the time taken by the algorithms to process millions of pixels or thousands of SIFT features in every frame.

## ACKNOWLEDGEMENT

This work was supported by the German Research Foundation (DFG) as part of the Transregional Collaborative Research Centre Invasive Computing (SFB/TR 89).